\documentclass{article}
\PassOptionsToPackage{numbers, compress}{natbib}
\usepackage[preprint]{neurips_2020}
\usepackage[utf8]{inputenc} 
\usepackage[T1]{fontenc}    
\usepackage{hyperref}       
\usepackage{url}            
\usepackage{booktabs}       
\usepackage{amsfonts}       
\usepackage{nicefrac}       
\usepackage{microtype}      
\usepackage{amsmath}
\usepackage{newunicodechar}
\usepackage{subcaption}
\usepackage{graphicx}
\usepackage{float}
\usepackage{hyperref}
\hypersetup{hidelinks}
\usepackage[noend]{algpseudocode}
\usepackage{algorithm}
\author{%
	Hilal AlQuabeh,  Ameera Bawazeer,  Abdulateef Alhashmi\\
	Department of Machine Learning\\
	Mohamed Bin Zayed University of Artificial Intellgence\\
	Abu Dhabi, UAE \\
	\texttt{20020117,20020018,20020034@mbzuai.ac.ae} \\
}

\usepackage{graphicx}
\title{Investigating a Baseline Of Self Supervised Learning Towards Reducing Labeling Costs For Image Classification}
\graphicspath{ {images/} }

\begin{document}
	
	\maketitle	
	\begin{abstract}
	Data labeling in supervised learning is considered an expensive and infeasible tool in some conditions.	The self-supervised learning method is proposed to tackle the learning effectiveness with fewer labeled data, however, there is a lack of confidence in the size of labeled data needed to achieve adequate results. This study aims to draw a baseline on the proportion of the labeled data that models can appreciate to yield competent accuracy when compared to training with additional labels. The study implements the kaggle.com' cats-vs-dogs dataset, Mnist and Fashion-Mnist to investigate the self-supervised learning task by implementing random rotations augmentation on the original datasets. To reveal the true effectiveness of the pretext process in self-supervised learning, the original dataset is divided into smaller batches, and learning is repeated on each batch with and without the pretext pre-training. Results show that the pretext process in the self-supervised learning improves the accuracy around 15\% in the downstream classification task when compared to the plain supervised learning. 	
\end{abstract}
\section{Introduction}
Image classiﬁcation is a very well-known machine learning problem in the computer vision domain. Image classiﬁcation task has been tested by different methods and techniques to provide state-of-the-art performance. Supervised learning via convolutional neural network provided generally the best accuracy given a huge annotated multiclass dataset.Huge labelled data are usually necessary to train deep neural network models to enhance visual feature learning for image classification. Self-supervised learning mechanism are suggested to learn overview image features from huge unlabelled data. Proxy labels will be simply created for the dataset in the pretext task, which will carry independent supervised task to train a model using the proxy labels. Therefore, the pre-trained model will boost the performance of the downstream image classification task.However, the success of a deep neural network highly relies on a large amount of labelled data used in training, which are often not available. Although large-scale datasets such as ImageNet, OpenImage, and CIFAR-10 allowed deep learning to achieve state-of-the-art performance, there are still way more unlabelled images on the internet that could increase the current state-of-the-art performance.In addition, labeling images for certain domains is very expensive, specially, for healthcare domain where labeling is done by professionals and Doctors. Therefore, reducing labeling cost is highly motivated by self supervised learning to utilize data without labels.Recent researches in self supervising learning focuses on the nature of datasets used in pretraining. For example, specialist pretraining, where models are pretrained on same dataset, outperformed generalist pretraining, where models are trained on similar yet different datasets. Another important dimension in SSL framework is the proportion of labeled and unlabeled data trained on to yield optimal performance.

\textbf{Cats vs. Dog Challenge:} 
Kaggle Cats vs Dogs challenge is popular challenge that challenge participants to develop algorithms to train a model that classify cats and dogs. Giving the public a common labelled cats and dogs dataset collected by Microsoft. Pierre Sermanet - New York, USA developed best classifier with 0.989 accuracy and won the first place along with thousands of submitted algorithms [1]. Therefore, Cats vs Dog Challenge became a well-known benchmark visual classification problem, where developers test new learning algorithms and models to validate their outcomes.

In this paper we deliver three main contributions:
\begin{itemize}
  \item Establish a baseline of SSL effectiveness in terms of the amount of labeled data
  \item Analyze pretext task accuracy and its relationship with downstream task accuracy 
  \item  Analyze how SSL by image rotation generalizes well to other datasets such as MNIST and Fashion-MNIST
  \end{itemize}

\subsection{Literature Review}
Considering the requirement of a deep neural network for large amounts of annotated data, it is necessary to come up with well-performing unsupervised learning techniques. Although fully unsupervised learning remains an unresolved challenge in computer vision research, the discovery of self-supervised learning opens a potential for better performance with a small amount of labelled training data. Before the discovery of self-supervised learning, several unsupervised visual feature representation methods were proposed. Singh et al. [2] proposed an unsupervised visual representation technique, which aims at discovering desired patches from a set of unlabelled images.Young et al. [3] proposed an algorithm that learns the features of ConvNet and clusters iteratively with the help of a recurrent framework. The framework comprises a clustering algorithm with successive operations that are expressed as a recurrent process. This clustering algorithm is placed on top of a representation output of a Convolutional Neural Network. The image cluster and representations are jointly updated during the forward pass of the training stage whereas representation learning is conducted in the backward pass. By doing so, the framework achieves a powerful representation and precise image cluster.

\begin{figure}[t]
	\centering
	\includegraphics[width=0.7\columnwidth]{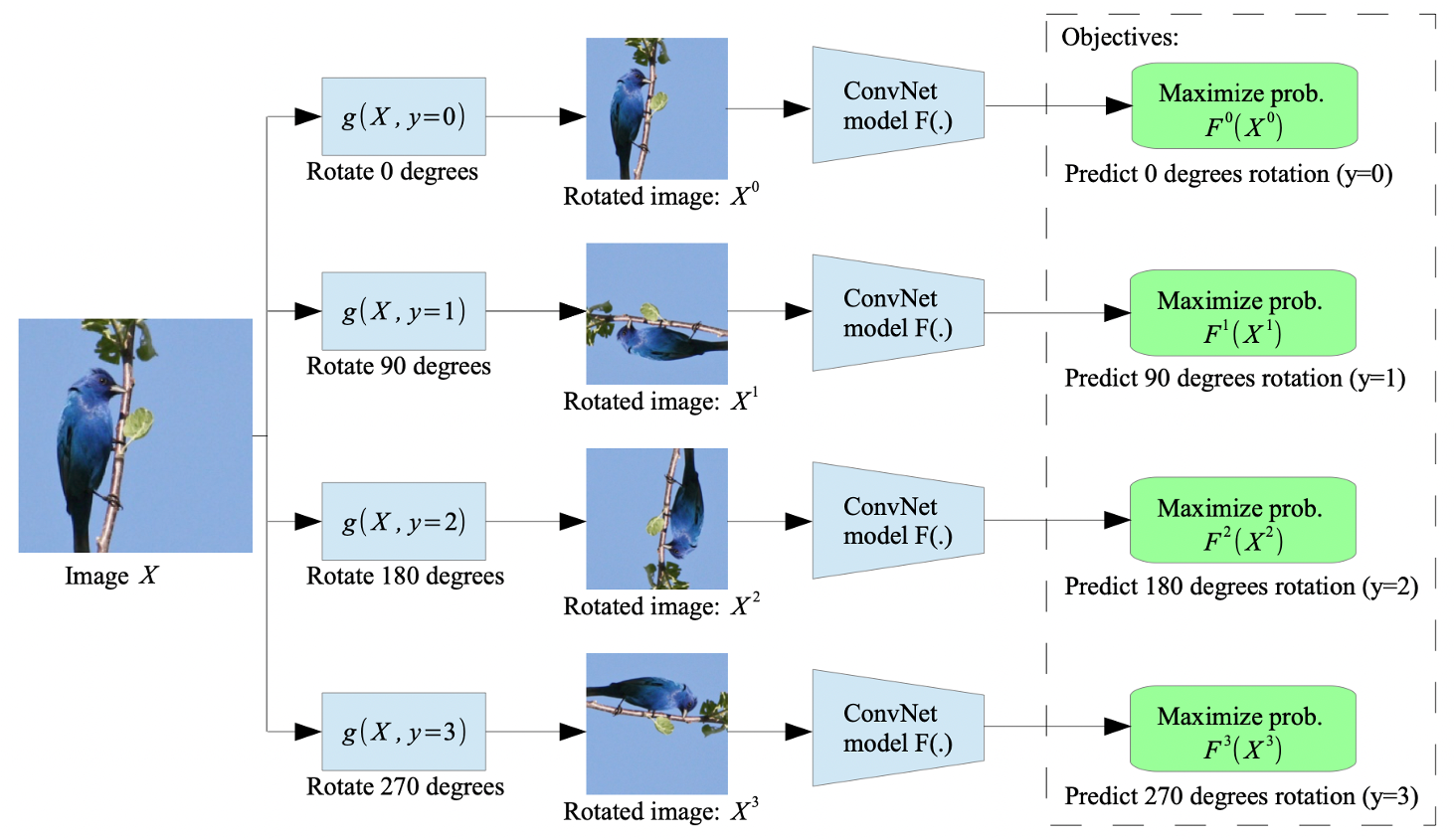}
	\caption{ Self-supervised task proposed for semantic feature learning by [8]}
	\label{fig:Spyros}
\end{figure}

Other augmentations were also successful in transferring valued features to the downstream task, such as jigsaw puzzle that predicts the correct arrangement of a jigsaw puzzle generated from the input image to learn the high-level image representations. Noroozi found that using jigsaw in self-supervised learn both the feature mapping and the proper spatial arrangement of object elements[4]. The scalability of self-supervision learning, specifically the Jigsaw and Colorization, is investigated by Priya Goyal et al. The aim of this paper is to see whether scaling the data, the model, or the problem complexity improves performance[5].
The two previous groups, in general, split input samples along a temporal line, predicting one provided the other. That could make it better suited to related tasks, such as semantic segmentation. According to Zhang et al, who used AlexNet to demonstrate comparative outcomes as compared to recent supervised methods, colorization is a pioneer of self-supervision [6]. The value of the failure, network design, and training information in achieving competitive results is demonstrated by Gustav Larsson's et al. thorough investigation into self-supervised colorization [7].
Spyros Gidaris et al. [8] demonstrated that predicting image rotation is a powerful method to train convolutional neural network to learn semantic features of input images without labels. Therefore, self-supervised feature learning enhances the performance of object detection, image classification and image segmentation considering limited amount of data labels. Spyros experimented image rotation in self-supervised learning setting in many benchmarks’ problems achieving state-of-the-art performance using different well-known datasets. Dosovitskiy et al. [9] apply ConvNet model training to achieve representations that are invariant on geometric transformations while discriminative between images. However, in image rotation pretext, a ConvNet model learn to recognize the geometric transformation applied to an image to predict one out of four [0, 90, 180, 270] image orientations as demonstrated in Figure \ref{fig:Spyros} . 
Unlike previous self-supervised approaches that rely on finite space of supervision labels and auxilary loss functions, Zhirong et al. [10] devised a novel self-supervised learning method in which each image instance is interpreted as a separate class, with a classifier trained to differentiate between individual input images. 
Nonetheless, no baseline has been established in the literature that can determine how many labelled data are required to achieve adequate image classification accuracy. This research aims to find such a baseline, if one exists, that can be applied to a variety of datasets and convolutional neural network architectures.

The following is the rest of the paper: The methodology of self-supervised learning, picture categorization, and model architecture are illustrated in section 2. Section 3 summarizes the findings of the ethe experiment and discusses the key foundations, while Section 4 wraps up the article with the most significant aspects.

	\section{Methodology}
\begin{figure}[h]
	\centering
	\includegraphics[scale=0.4]{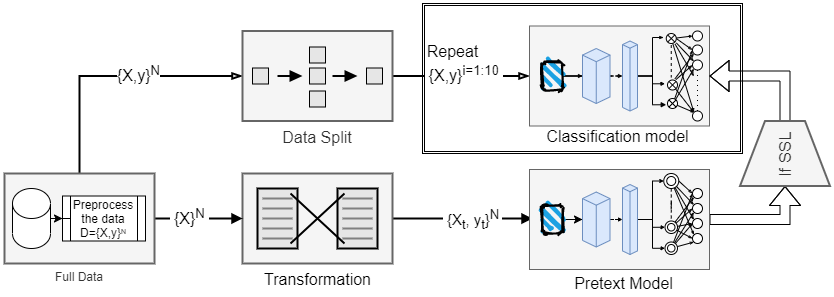}
	\caption{The experiment flow begins with preprocessing the data, followed by applying a transformation to the images while simultaneously generating appropriate labels, pretraining a CNN classifier to predict the applied transformation, and finally transferring the learned representations to the downstream task.}
	\label{fig:flow}
\end{figure}

In order to achieve the aim of this study, which is to establish a baseline of SSL effectiveness in terms of the amount of labeled data, a simple yet promising workflow is designed and implemented as shown in figure \ref{fig:flow}.The self-supervised effects on task performance are directly contrasted to the case of supervised learning using the dogs-cats dataset from kaggle and other datasets, where the self-supervised influence on task (image classification) performance are the main goal compared to the conventional supervised scenario. In particular, the amount of data that is labelled and available for the downstream task, which is the main classification task, is varied from a turning level to the maximum quantity of data in order to determine the relationship or benchmark of this increase if exist.\par 
\subsection{ResNet Archeticture}
\begin{figure}[b]
	\centering
	\includegraphics[width=0.8\columnwidth]{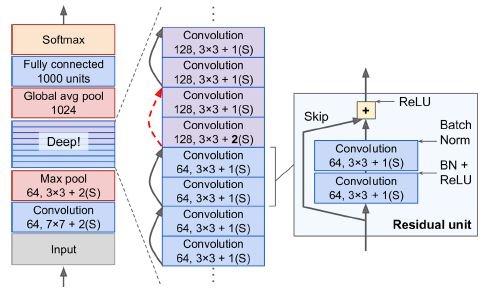}
	\caption{ResNet Archetecture [11]}
	\label{fig:resnet}
\end{figure}

The Residual Network (ResNet) developed by Kaiming He et al. [11] has an extremely deep CNN (up to 152) layer while keeping the number of parameters modest. The use of skip connections and the signal feeding into a layer being added to the output of a layer deeper upstream is the key trick to being able to train such a deep network. The ResNet architecture has proven to be effective in image classification and a variety of other computer vision tasks. The ResNet architecture with 20 layers is incorporated into the pretext and downstream tasks in this paper study. Figure \ref{fig:resnet} illustrates the main layers used in ResNet 20. 

\subsection{Pretext Task}
To compensate for the lack of labeled data, the pretext method in SSL applies a rotational transformation to the available inputs while simultaneously labeling them with a proxy label. The new pseudo labelled are four correspond to four angles of rotations (0 90, 180 and 270). The generted data can be thought of as a new classification task, in which training a convolutional neural network (CNN) using ResNet archetizture with twenty layers (ResNet-20) to predict the pseudo label can benefit in learning the feature representations of the data. \par The task's objective is to minimize the cross entropy losses incurred by predicting the four output neurons against the ground truth in the pdesuo labels. If we denote the rotation applied to an image by $T(.|y^t)$ where $y^t$ is a binary variable activiate for every rotation angle \textit{t} (four options) then a set of four rotations  (classes) is created that the model has to predict from $R=\{T(.|y^t)\}_{t=1}^{4}$. The classifier so generate a probablitc output among all possible classes $M(X|\theta)=\{M^{t}(X|\theta)\}_{t=1}^{4}$ and so the loss of single example become as shown in equation 1.
\begin{align}
L = -\sum_{t=1}^4  y^tlog(M^{t}(X|\theta))
\end{align}

\subsection{Downstream task-Image Classification}

The question that this paper attempts to address is how much unlabeled data is needed to learn a visual representation that can compete with standard supervised learning with equivalent label data in the downstream task (image classification). The response is discovered by running two simultaneous training sessions on ten separate data volumes, one with SSL and the other without as illustrated in algorithm 1 . After the data is divided into ten stratified kfolds, and the algorithm feeds it to the pre-trained model resulted from the pretext task, with the last two layers fine-tuned. Simultaneously, the same data volume is fed to a ResNet model that has not been trained as illustrated by figure 2. THe resulted accuracy of image classifier is compared in both scenarios to draw a conclusion about the number of labelled data influencing the downstream task.
\begin{algorithm}
	\caption{Parallel training with and without self-supervised pretext training}
	\begin{algorithmic}[1]
		\Procedure{SSL}{$\{X_k,y_k\}_{k=1}^N$}       
		\State \textbf{Transformation of input data}
		\State $T(.|y^t) : \{X_k\}_{k=1}^N\rightarrow \{X_k^t\}_{k=1}^N$ 
		\Comment{$y^t=\{0,90,180,270\}$}
		\State \textbf{Pretext model}
		\State $M(\cdot|\theta) : \{X^t\}^N\rightarrow \{X^t,y^t\}^N$
		\State Downstream model initialization
		\State $\{X,y\}^N\rightarrow \{X,y\}^{i}$ where $i=1,...,10$ and $\sum\{\cdot\}^{i}=\{\cdot\}^N$
		\For{$i=1:10$} 
		\If{$i > 10$}
		\State Break
		\ElsIf{$i > 1$}
		\State $\{X,y\}^i=\{X,y\}^{i-1}+\{X,y\}^{i}$
		\Else
		\State $\{X,y\}^i=\{X,y\}^{i}$
		\EndIf
		\If{with SSL}
		\State Call pre-trained model
		\Comment{Apply fine-tunning }
		\State	$sSL(\cdot|\lambda) : \{X\}^i\rightarrow \{X,y\}^i$
		\Else
		\State
		$SL(\cdot|\lambda) : \{X\}^i\rightarrow \{X,y\}^i$
		\EndIf
		\EndFor
		\Return $SL(\cdot|\lambda)$ , $sSL(\cdot|\lambda)$
		\EndProcedure
	\end{algorithmic}
\end{algorithm}

\subsection{Representation Learning}
\begin{figure}[t]
	\begin{subfigure}{.5\textwidth}
		\centering
		\includegraphics[width=.8\linewidth]{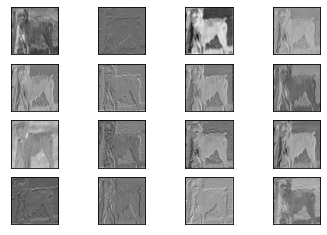}  
		\caption{Classification task model}
		\label{fig:fm1}
	\end{subfigure}
	\begin{subfigure}{.5\textwidth}
		\centering
		\includegraphics[width=.8\linewidth]{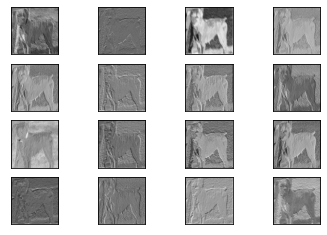}  
		\caption{Pretext task model}
		\label{fig:fm2}
	\end{subfigure}
	\caption{Feature maps generated by two models: pretext and the conventional image classification using ResNet-20}
	\label{fig:fm}
\end{figure}

The basic concept behind using image rotations as a pre-training task before the main classification task is that any CNN model trained on any task would attempt to identify and become familiar with the data features and semantic parts in images that lead to the representation acquired. In order to predict effectively, the pretext model must relate the sections of images, hidden features, and the orientation of its background. Figure \ref{fig:fm1} shows the feature maps created by the conventional classification model without the pre-training component. Despite the fact that the pretext model's goal is rotation predictions and not the background component, the feature maps in figure \ref{fig:fm2} are produced by the pretext model trained on the rotation recognition task, where they appear to be similar to the attention maps in figure \ref{fig:fm1} if we assume high density corresponds to the main parts of the object being identified.

\section{Results}

In our results section we conduct an extensive evaluation of self supervised learning task and downstream task implementation on Cats vs Dogs dataset [12].  We do this by comparing the quality of the learned SSL model with respect to the supervised learning scenario. We study the effect of size of labeled data on the performance of both SSL and SL models. Further A correlation analysis is performed to study whether correct learning in the SSL task influence the performance of downstream task. Finally, we test the generality of our approach with other commonly used image datasets such as MNIST [13]  and Fashion MNIST [14] . 

\subsection{Experiment on Cats vs Dogs classification}

\subsubsection{Data augmentation and preprocessing}
The Cats and Dogs dataset is downloaded from TensorFlow classification datasets. We first change the type of the data from prefetch to NumPy to be able to process it using TensorFlow functions. The dataset images have a non-unified shape, so we start by unifying the shape of all images to 120x120x3. 120x120 was chosen by trial and error and by benchmarking other existing cats and dogs classification projects. The pixel values are normalized by dividing the pixel values by 255, this puts them on the same scale between 0 to 1.\par 
Random width and height shift is applied by 1\%, moving all pixels of the image in one direction, such as horizontally or vertically, while keeping the image dimensions the same. There will be a region of the image where new pixel values will have to be specified, these are filled based on the nearest pixel value. 

\subsubsection{Self supervised learning task }
A rotation augmentation proxy task is designed such that the training and test images are randomly rotated to 0,90,180, and 270 degrees. A new label for each rotated image is generated indicating the degree of rotation of each image. This now becomes a set of rotated images with their labels. We treat this task as a supervised learning problem where the goal is to output the degree of rotation for each input image.\par

To train a model for this classification task, we choose ResNet deep neural network [15] with depth 20. ResNet model architecture is kept the same except removing the last layer of the model substitute it with a densely connected NN layer with SoftMax activation. The final layer outputs 4 categories now. The model is trained in 200 epochs, each epoch takes in images in batches of 128 images. We compile the model using Adam as our optimizer and categorical cross-entropy as our loss function. Adam [16] is a stochastic gradient method used with its default parameter values for learning rate, exponential decay rates, and epsilon. The metric used is the model accuracy indicating how often predictions equal the ground truth labels. Categorical cross-entropy loss is used to determine how well our neural network architecture fits the data, and is measure by equation 1. 
\begin{equation}
CE = -\log\left(\frac{e^{S_p}}{\sum_{j}^{C}e^{S_j}}\right)
\end{equation}
\begin{figure}[t]
	\begin{subfigure}{.5\textwidth}
		\centering
		\includegraphics[width=\linewidth]{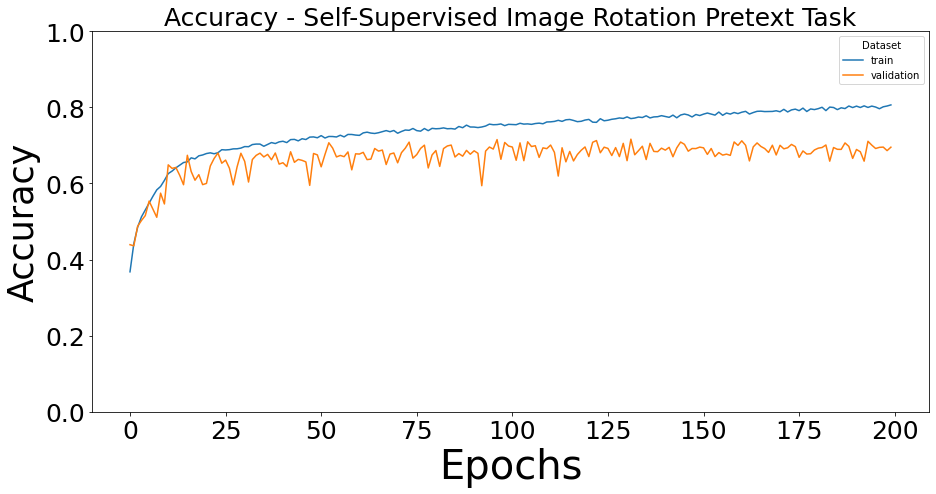}  
		\caption{Training and validation accuracy }
		\label{fig:AccSLL}
	\end{subfigure}
	\begin{subfigure}{.5\textwidth}
		\centering
		\includegraphics[width=\linewidth]{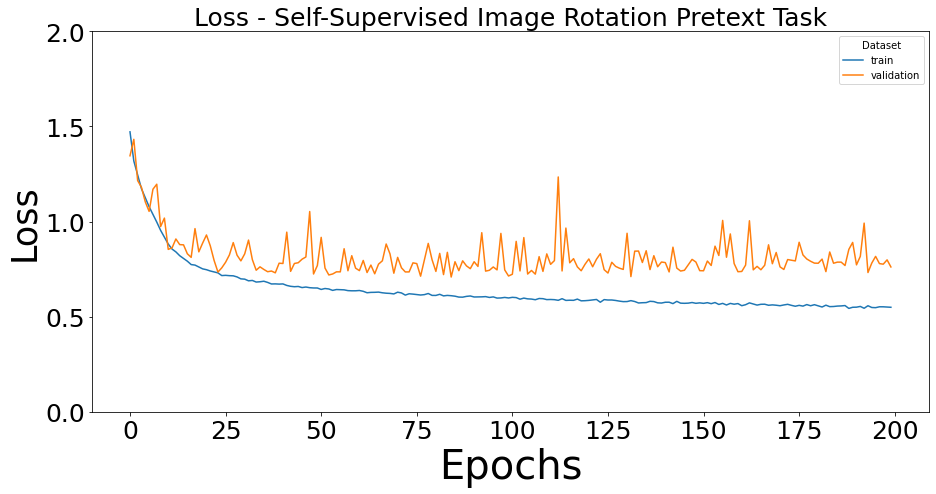}  
		\caption{Training and validation loss }
		\label{fig:LossSLL}
	\end{subfigure}
	\caption{Comparison of cats and dogs images for the SSL rotation task over the 200 epochs.}
	\label{fig:pretext1}
\end{figure}
Where $S_p$ is the network score for the positive class and $ S_j$ is the score inferred by the net for each class C. The model is fitted with the training data and validated with the validation data. The models are then saved to be used later in the downstream task.  In Figures \ref{fig:AccSLL}and  \ref{fig:LossSLL} we provide a comparison between the training and validation accuracy and loss over increasing epochs number. The experiment is set to run for 200 epochs. The accuracy increases exponentially during the first 25 epochs reaching 0.6910 for the training set and 0.6614 for the validation set. After that the increase in accuracy is relatively slower. At the end of the 200 epochs the accuracy of the rotation classification task reaches 0.8106 for the training set and  0.6954 for the validation set. A similar trend is seen in the training loss. The training loss at epoch 25 for the training and validation 0.7133 and 0.7860 respectively. A slower decrease in loss is then observed to finalize at 0.5416 and 0.7613 in the 200 epoch.

\subsubsection{Downstream classification task}

In the downstream task, we train the SSL model using the pre-trained ResNet20 from the pretext task. We compile the model using the same optimizer, loss function, and metric as the pretext task and we run it for 200 epochs. However, we use the ResNet scheduled learning rate so that the learning rate is scheduled to be reduced after 80, 120, 160, 180 epochs. We also implement the technique of reducing the learning rate when accuracy  stop improving after 5 successive epochs. The learning rate is reduced by a square root of 0.1, but we bound it by a minimum learning rate of 0.5e-6.Since Keras workers can make experiments much faster, we set workers to 4 when fitting the model. This can help in speeding the converge rate for all the runs, however, special care should be given to the learning rate.\par

We investigate the effect of the number of labeled input data on the accuracy and loss of the self-supervised model comparing it to a supervised model of same ResNet20 architecture initialized with random weights . To do this we start by using stratified k-fold cross validation of 50 folds splitting our 18,610 training data into 18,238 train and 372 test data points approximately. Stratified k-fold ensures that data increments consist of class balanced data.We fit the supervised and self supervised models by using the test set from the stratified k-folds split originating from the actual training set of our images. In each of the 10 models, we increase the input size by 372 samples, so that in model 10 the input size is  around 3,722 images. The models are validated with the actual validation set of size 4,652 outside the 18,238 training set. Finally the best accuracy and loss performances are recorded per model and saved . 
\begin{figure}[h]
	\begin{subfigure}{.5\textwidth}
		\centering
		\includegraphics[width=\linewidth]{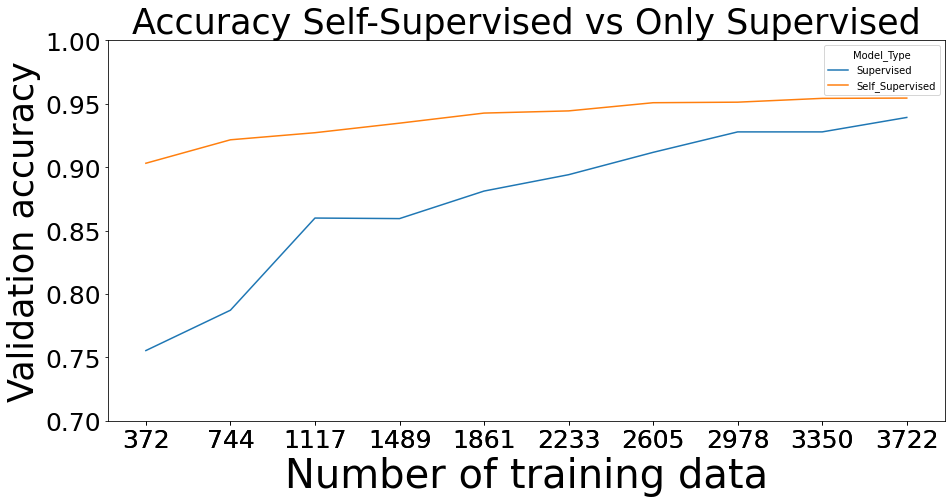}  
		\caption{Self supervised and supervised validation accuracy }
		\label{fig:Accdown}
	\end{subfigure}
	\begin{subfigure}{.5\textwidth}
		\centering
		\includegraphics[width=\linewidth]{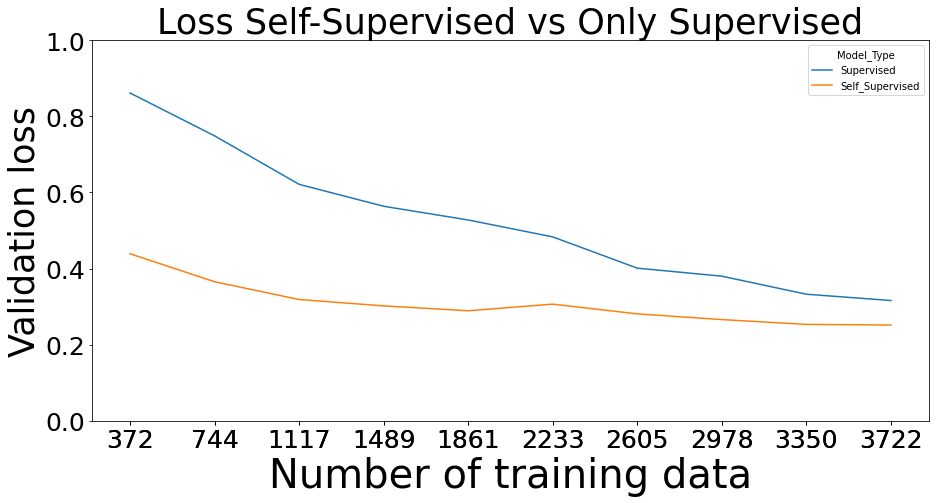}  
		\caption{Self supervised and supervised validation loss }
		\label{fig:Lossdown}
	\end{subfigure}
	\caption{Cats-dogs classification results for supervised and self-supervised learning }
	\label{fig:down}
\end{figure}

Figures \ref{fig:Accdown} and \ref{fig:Lossdown}  show the validation accuracy and losses repeated at every labeled data split for the downstream process. This is done for both scenarios,  with and without the transferred learning from the pretext process.
The results clearly show improve in performance when the self supervision task is implemented. Moreover, the performance improvement gap is larger with less labeled data. We can see from figure \ref{fig:Accdown} when the models are trained with only 372 labelled training set (2\% of training set),  accuracy difference is almost around 14\%. However, as the labeled data increase the performance of supervised model improves and the improvement of self supervised model become slow. For instance, when using 3722 labeled images the accuracy of SSL model is 0.954 while the supervised model accuracy is 0.939. These accuracies are almost similar, and the gain from self supervision is only 0.015.\par

\begin{figure}[t]
	\begin{subfigure}{.5\textwidth}
		\centering
		\includegraphics[width=\linewidth]{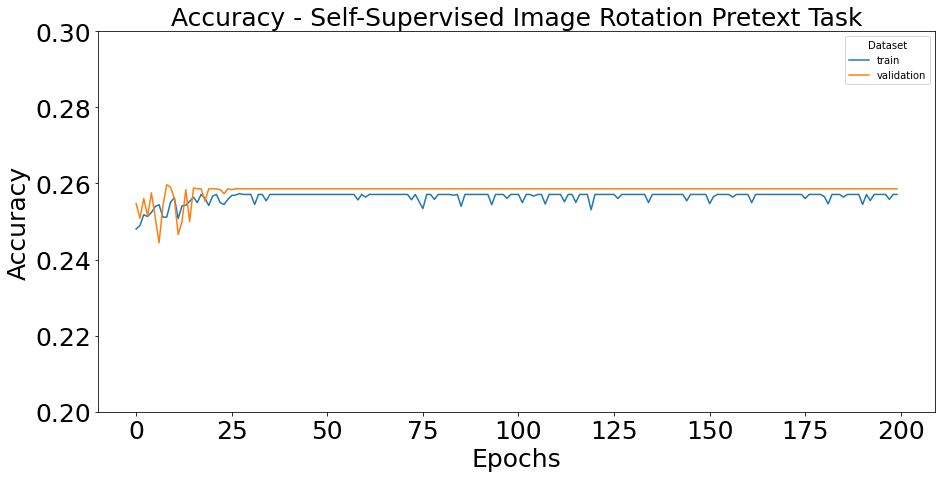}  
	         \caption{Training and validation accuracy  }
\label{fig:randupacc}
	\end{subfigure}
	\begin{subfigure}{.5\textwidth}
		\centering
		\includegraphics[width=\linewidth]{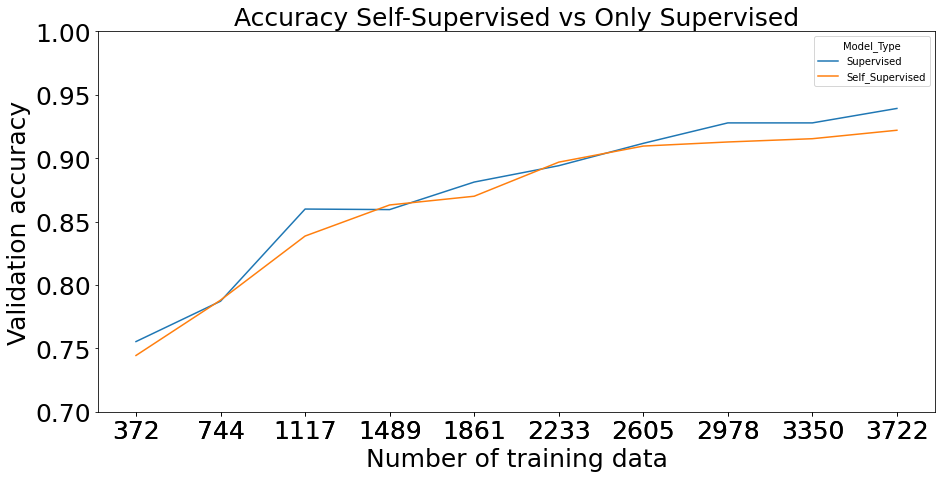}  
	         \caption{Self supervised and supervised validation accuracy }
\label{fig:randownacc}
	\end{subfigure}
	\caption{Cats-dogs classification results for supervised and self-supervised learning using inconsistent self supervised learning }
	\label{fig:down}
\end{figure}

\textbf{Investigating the influence of pretext task on the downstream classification:} we study the relation between the accuracy of the pretext task rotation on the downstream task. We conduct this experiment on the same cats and dogs dataset with the same experiment design. However, in this experiment we try to randomize the ground truth label for each rotation. The random rotation labelling assigns inconsistent label to each degree of rotation, so that an image that is rotated 90 degrees will have a ground truth label of 90, 180, or 270 arbitrary. As expected this result in low performance in the pretext task learning. As shown in figure \ref{fig:randupacc} the SSL task accuracy is very low for both the training and validation set. This is also reflected in the downstream classification task as shown in figure \ref{fig:randownacc}. The figure suggests that there is no clear gain on the accuracy of self supervised model compared to the supervised model on the classification task.

\begin{figure}[b]
	\begin{subfigure}{.5\textwidth}
		\centering
		\includegraphics[width=\linewidth]{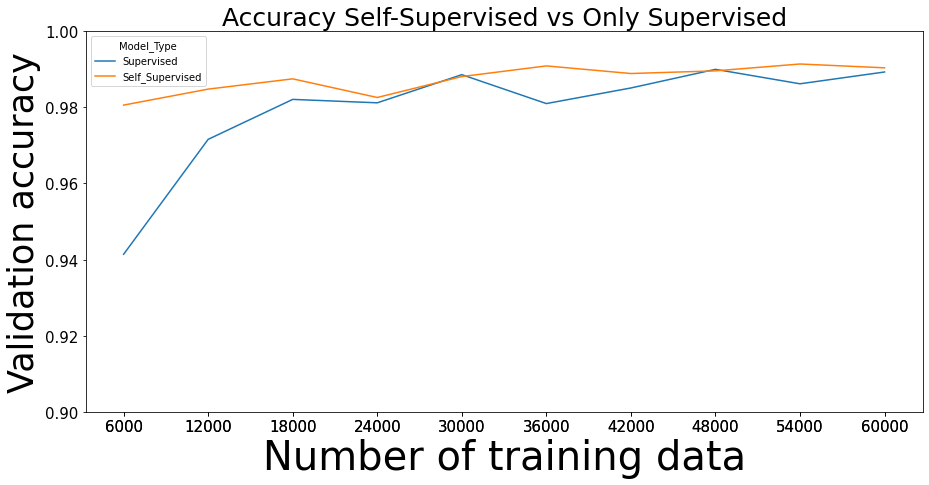}  
		\caption{Self supervised  and supervised validation accuracy}
		\label{fig:mnsitacc}
	\end{subfigure}
	\begin{subfigure}{.5\textwidth}
		\centering
		\includegraphics[width=\linewidth]{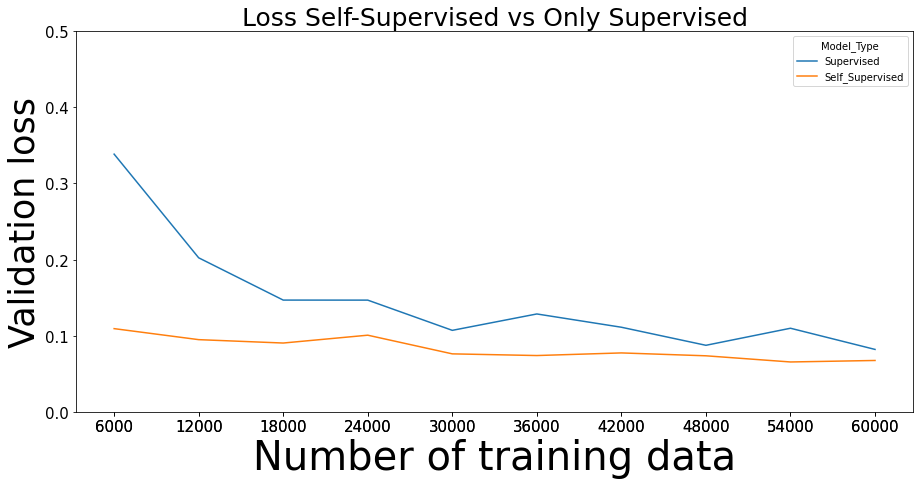}  
		\caption{Self supervised and supervised validation loss }
		\label{fig:mnsitloss}
	\end{subfigure}
	\caption{MNIST classification results for supervised and self-supervised learning }
	\label{fig:downmnsit}
\end{figure}

\subsection{Experiment on other datasets}

\textbf{Task generalizability on MNIST data:} We further explore the generalizability of the framework with MNIST dataset, consisting of 60,000 training images and 10,000 test images of size 28x28x1. We use ResNet20 and follow exactly the same design used for Cats and dogs experiment. We fix all parameters and remove the last layer of the model substitute it with a densely connected NN layer with SoftMax activation. The model now performs classification for 10 categories instead of 4.  In the downstream we compare the SSL model with the supervised model using stratified k-fold cross validation of k = 10 and then we choose to create 10 models. Each of the 10 models will be trained on training data starting from from 6,000 labelled input for model 1 to 60,000 labelled input for model 10. As shown from Figure \ref{fig:downmnsit} the accuracy gap between the supervised and self supervised model relativly smaller compared to the cats and dogs data. \par Following the same trend from cats and dogs experiment, the highest advantage from self supervision is gained when the both models are trained on lowest fraction of data. Here 6,000 images only (10\% of training data) provided an advantage for an increase in classification accuracy for SSL with a gap of around 4\%. When provided with full data, the SSL downstream model accuracy the supervised model accuracy is almost similar with 0.990 and 0.989 respectively.

\begin{figure}[t]
	\begin{subfigure}{.5\textwidth}
		\centering
		\includegraphics[width=\linewidth]{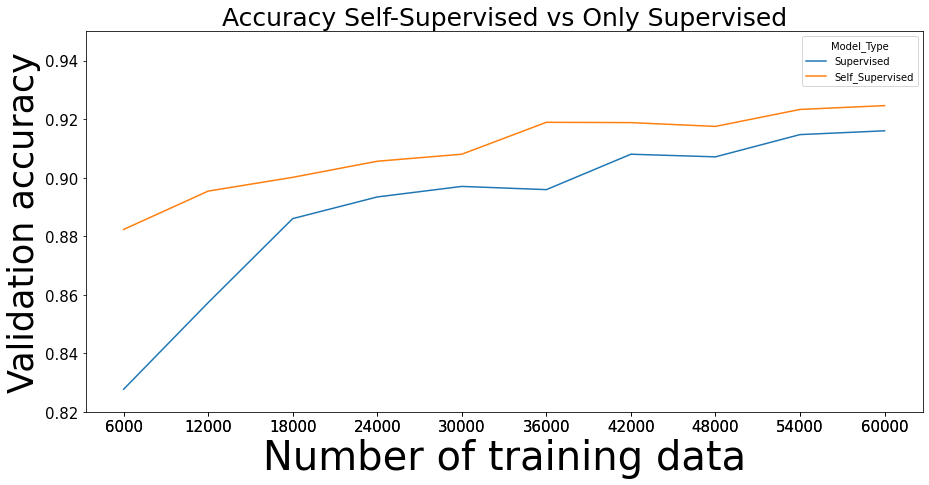}  
	         \caption{Self supervised  and supervised validation accuracy }
\label{fig:fashionacc}
	\end{subfigure}
	\begin{subfigure}{.5\textwidth}
		\centering
		\includegraphics[width=\linewidth]{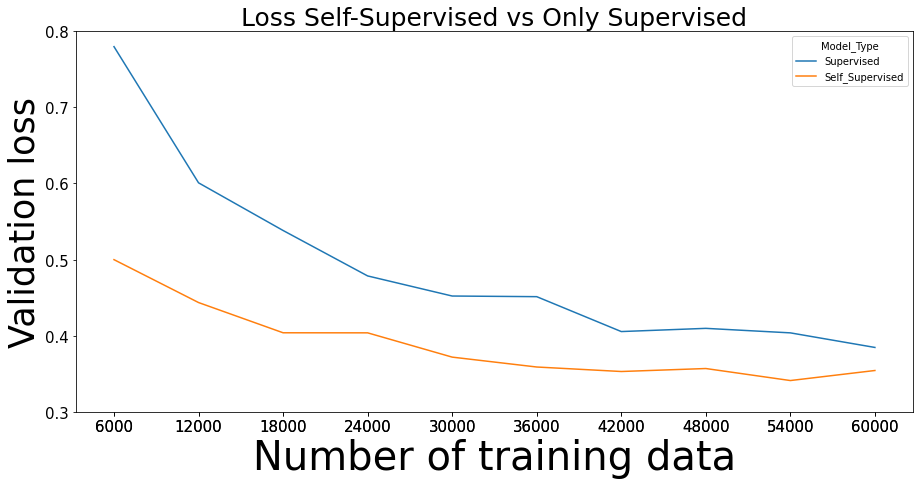}  
	         \caption{Self supervised and supervised validation loss}
\label{fig: fashionloss}
	\end{subfigure}
	\caption{Fashion-MNIST classification results for supervised and self-supervised learning}
	\label{fig:downfashmnsit}
\end{figure}

\textbf{Task generalizability on Fashion-MNIST data:} In Fashion-MNIST we repeat the same experiment done for the MNIST dataset. The training set consists of 60,000 grey scale images and 10,000 test images of size 28x28x1. The downstream model performs image classification for 10 classes similar to the MNIST classification. As shown from Figure \ref{fig:fashionacc} the accuracy gap between the supervised and self supervised model is largest for smaller fraction of training data. Similar to the results of the previous experiments, using 6,000 images only (10\% of training data) provided an advantage for an increase in classification accuracy for SSL with a gap of around 0.05.

\textbf{Discussion:} In Table \ref{table:comp}  we summarize the results for our experiments in a single table showing how different datasets benefit  from different fraction of labelled training data. For each dataset, we see how much gain the self supervision provides to the task accuracy given small fraction of labelled training data (i.e., 2\% and 10\%).  

\begin{table}[ht]
\centering 
\caption{ Benefit of each dataset from fraction of labelled training data  } 

\begin{tabular}{p{2cm} p{1.5cm} p{3cm} p{1.5cm} p{3cm}} 
	\toprule
 Dataset & Fraction of training data & Accuracy gain over supervised network\% & Fraction of training data & Accuracy gain over supervised network\%  \\ [0.1ex] 
\hline 

Cats vs Dogs & 10\% & 6.15 & 2\% & 14.80 \\ 
\hline 
Fashion-MNIST & 10\% & 5.46 & 2\% & 10.1  \\ 
\hline 
MNIST & 10\% & 4.00 & 2\%& 4.30 \\ [0.8ex]

\end{tabular}

\label{table:comp} 
\end{table}

\subsection{Conclusion}

Self-supervised learning allows deep neural network model to be trained utilizing less amount of labelled data. Random rotation pretext task allowed observable sematic feature learning for input images without labels. These learned visual features boosted the downstream image classification task remarkably. This paper attempts to draw performance baseline between SL and SSL models based on amount of labelled data trained on. Results showed that SSL outperformed SL in image classification task when both models are trained on different sizes of labelled data. Highest advantage of downstream task accuracy is gained from training with low fraction of labelled data, which highlight the advantage of SSL dealing with less labelled data. Extending the experiment to Fashion and MNIST datasets, SSL by image rotation generalizes very well to other datasets, and highest gain is noticed with more challenging datasets. Further experiments to random rotation pretext showed that high pretext task accuracy in SSL, resulted in more semantic feature learning, which translate to higher downstream task accuracy.

\section*{References}
\small
[1]kaggle. 2013. cats-vs-dogs. Retrieved 4/5/2021 from kaggle.com/c/dogs-vs-cats.\par
[2] S. Singh, A. Gupta, and A. A. Efros, “Unsupervised discovery of mid-level discriminative patches,” in European Conference on Computer Vision. Springer, 2012, pp. 73–86. 2\par
[3] J. Yang, D. Parikh, and D. Batra, “Joint unsupervised learning of deep representations and image clusters,” in Proceedings of the IEEE conference on computer vision and pattern recognition, 2016, pp. 5147–5156. 2\par
[4] M. Noroozi and P. Favaro, “Unsupervised learning of visual representations by solving jigsaw puzzles,” in European conference on computer vision.Springer, 2016, pp. 69–84. 3\par
[5] P. Goyal, D. Mahajan, A. Gupta, and I. Misra, “Scaling and benchmarking self-supervised visual representation learning,” in Proceedings of the IEEE/CVF International Conference on Computer Vision, 2019, pp. 6391–6400. 3\par
[6] R. Zhang, P. Isola, and A. A. Efros, “Colorful image colorization,” in European conference on computer vision. Springer, 2016, pp. 649–666. 3\par
[7] G. Larsson, M. Maire, and G. Shakhnarovich, “Colorization as a proxy task for visual understanding,” in Proceedings of the IEEE Conference on Computer Vision and Pattern Recognition, 2017, pp. 6874–6883. 3\par
[8] S. Gidaris, P. Singh, and N. Komodakis, “Unsupervised representation learning by predicting image rotations,” arXiv preprint arXiv:1803.07728, 2018.2\par
[9] 
[10] Z. Wu, Y. Xiong, S. X. Yu, and D. Lin, “Unsupervised feature learning via non-parametric instance discrimination,” in Proceedings of the IEEE Conference
on Computer Vision and Pattern Recognition, 2018, pp. 3733–3742.3\par
[11] He, Kaiming, Xiangyu Zhang, Shaoqing Ren, and Jian Sun. "Deep residual learning for image recognition." In Proceedings of the IEEE conference on computer vision and pattern recognition, pp. 770-778. 2016.\par
[12] Elson, Jeremy, Douceur, John (JD), Howell, Jon, Saul, Jared., “Asirra: A CAPTCHA that Exploits Interest-Aligned Manual Image Categorization”, Proceedings of 14th ACM Conference on Computer and Communications Security (CCS), 2007, (https://www.microsoft.com/en-us/research/publication/asirra-a-captcha-that-exploits-interest-aligned-manual-image-categorization/)\par
[13] Y. LeCun and C. Cortes, “MNIST handwritten digit database,” 2010. \par
[14] Han Xiao and Kashif Rasul and Roland Vollgraf, “Fashion-MNIST: a Novel Image Dataset for Benchmarking Machine Learning Algorithms”, 2017, (https://arxiv.org/abs/1708.07747)\par
[15] K. He, X. Zhang, S. Ren, and J. Sun, “Deep Residual Learning for Image Recognition,” presented at the 2016 IEEE Conference on Computer Vision and Pattern Recognition (CVPR), Jun. 2016, doi: 10.1109/cvpr.2016.90.\par
[16] Kingma, Diederick P and Ba, Jimmy, Adam: A method for stochastic optimization, International Conference on Learning Representations (ICLR), 2015.

\medskip

\end{document}